# Brain Computer Interface for Gesture Control of a Social Robot: an Offline Study

**Reza Abiri**
Department of Mechanical, Aerospace, and Biomedical Engineering
University of Tennessee, Knoxville
Knoxville, TN 37996, USA
rabiri@vols.utk.edu

**Griffin Heise**
Department of Mechanical, Aerospace, and Biomedical Engineering
University of Tennessee, Knoxville
Knoxville, TN 37996, USA
gheise@vols.utk.edu

**Xiaopeng Zhao**
Department of Mechanical, Aerospace, and Biomedical Engineering
University of Tennessee, Knoxville
Knoxville, TN 37996, USA
xzhao9@utk.edu

**Yang Jiang**
Department of Behavioral Science
College of Medicine
University of Kentucky
Lexington, KY 40356, USA
yjiang@uky.edu

**Fateme Abiri**
Department of Computer Engineering
Ferdowsi University of Mashad
Mashad, Iran
fateme.abiri@yahoo.com

*Abstract*— Brain computer interface (BCI) provides promising applications in neuroprosthesis and neurorehabilitation by controlling computers and robotic devices based on the patient's intentions. Here, we have developed a novel BCI platform that controls a personalized social robot using noninvasively acquired brain signals. Scalp electroencephalogram (EEG) signals are collected from a user in real-time during tasks of imaginary movements. The imagined body kinematics are decoded using a regression model to calculate the user-intended velocity. Then, the decoded kinematic information is mapped to control the gestures of a social robot. The platform here may be utilized as a human-robot-interaction framework by combining with neurofeedback mechanisms to enhance the cognitive capability of persons with dementia.

## I. INTRODUCTION

The field of human-robot interaction has been significantly enriched with the integration of Brain computer interface (BCI), in which the subject can manipulate the environment in a desired way compatible with his/her intention through the brain activities [1]. For example, integrating BCI into exoskeletons, rehabilitation robots, and prosthetics has shown increased efficiency of rehabilitation due to direct intention of patient in rehabilitation progress [2-8].

In BCI and particularly in noninvasive approaches, electroencephalography (EEG) based paradigms are more convenient and portable than other neuroimaging techniques such as electrocorticography (ECoG), magnetoencephalography (MEG), and magnetic resonance imaging (MRI) [1]. Many different EEG paradigms have been developed using external stimulations, sensorimotor rhythms, or imaginary motor movements. The main drawback for systems on sensorimotor rhythms is the lengthy training time (several weeks to several months) required for the subjects to achieve satisfactory performance. In cases with external stimulations, a fatigue phenomenon has been reported by subjects and researchers, although it should be noted that this paradigm is not reflecting the user's intention to control a device. Another issue concerning these paradigms is the discrete control of cursor directions due to switching among imagined movements of several large body parts [9] or switching among multiple paradigms [10]. The alternative system based on imaginary movement, as first designed by Bradberry et al. [11], has the capability to minimize the training time (~20 minutes for two dimensional cursor control).

Many researchers have employed EEG paradigms to control robotic systems. Sensorimotor rhythms have been utilized by various authors to control remote robotic systems [12], virtual and real quadcopters [13-15], and robotic arms [16-18]. Using an external stimulation paradigm/hybrid paradigm, researchers demonstrated the control of a prosthetic arm [19], artificial arm [20], and an exoskeleton for rehabilitation of the hand [21]. Besides the brain-controlled robots such as mobile robots [22], controlling humanoid and social robots has become of interest in BCI. Social robots are autonomous robots that can interact and communicate with humans. For example, by employing the aforementioned EEG paradigms, some researchers controlled the movements of humanoid robots such as NAO through direct control approaches [23-28]. However, no previous work had been reported on manipulating a humanoid/social robot in cognitive training for patients with cognitive deficits. In this work, we develop a novel neurofeedback-based noninvasive BCI system for possible applications in cognitive enhancement. In contrast to previous studies on computer-based neurofeedback systems, the platform here is based on interaction with a social robot. The interaction with a robot may better engage and motivate user participation in specified tasks and thus enhance the targeted rehabilitation program. An initial testing of the developed platform is conducted using the imagined body kinematics scheme originally proposed in [11] to control different gestures of a social robot.

## II. MATERIALS AND METHODS

*A. Experimental protocol*

Before controlling the social robot, the subjects are instructed to use a BCI system in a cursor control task. The experiment served two objectives; first, a regression model was developed to extract imagined body kinematics from the subject's brainwaves. Second, the experiment helped to familiarize the subject with BCI concepts. The institutional Review Board of the University of Tennessee approved the experimental procedure and 5 subjects (4 male, 1 female) participated in the experiments after signing the informed consent. For the experiments, a PC with dual monitor was provided. One monitor for the experimenter and another for the subjects.

During the experiments, EEG signals were acquired by using an Emotiv EPOC device with 14 channels and through BCI2000 software (with 128 sampling time, high pass filter at 0.16Hz, and low pass filter at 30Hz). The cursor control task included three phases. Phase 1 was the training phase. The subject was asked to sit comfortably in a fixed chair with hands resting in the lap. The subject's face was kept at an arm's length from the monitor. The subject was instructed to track the movement (up-down/right-left) of a computer cursor, whose movement was controlled by a practitioner in a random manner. Meanwhile, the subjects were instructed to imagine the same matched velocity movement with their right index fingers. The training phase consisted of 5 trials, each of which lasted 60 seconds. Phase 2 was the calibration phase, during which a decoder model was constructed to model the velocity of the cursor as a function of the EEG waves of the subject. For more accurate reconstruction and prediction of the imagined kinematics at each point, 5 previous points (time lag) of EEG data were also included in the decoding procedure. Then, the developed decoder was fed into BCI2000 software to test the performance of the subject in phase 3 (test phase). In the test phase, the subject was asked to move the cursor using their imagination to a target that randomly appeared at the edges of the monitor.

*B. Decoding*

Many decoding methods for EEG data have been investigated by researchers in the frequency and time domains. Most sensorimotor-rhythms-based studies are developed in the frequency domain [9, 13-15, 17, 18, 29-32]. Meanwhile, in the time domain, researchers employed regression models as a common decoding method for decoding EEG data for offline decoding [33-37] and real-time implementation [11]. Some nonlinear methods such as the Kalman filter [38] and the particle filter model [39] were also applied in decoding EEG signals for offline analysis. Many previous works confirmed that among kinematics parameters (position, velocity), velocity encoding/decoding shows the most promising and satisfactory validation in prediction [33, 34, 36]. Hence, we were motivated to decode and map the acquired EEG data to the observed velocities in x and y directions. In other words, the aim was to reconstruct the subject's trajectories off-line from EEG data and obtain a calibrated decoder. For this purpose, all the collected data was transferred to MATLAB software for analyzing and developing a decoder. Here, based on a regression model for output velocities at time sample $t$ in x direction ($u[t]$) and y direction ($v[t]$), the equations can be presented as follows:

$$u[t] = a_{0x} + \sum_{n=1}^{N}\sum_{k=0}^{K} b_{nkx} e_n[t-k] \qquad (1)$$

$$v[t] = a_{0y} + \sum_{n=1}^{N}\sum_{k=0}^{K} b_{nky} e_n[t-k] \qquad (2)$$

where $e_n[t-k]$ is the measured voltage for EEG electrode $n$ at time lag $k$ and for the total number of EEG sensors $N=14$ and total lag number $K=5$. Based on a previously published study [11], for more accurate reconstruction of the imagined kinematics, 5 previous points (time lag) of EEG data were included in the decoding and prediction of present value. The choice of 5 lag points is the tradeoff between accuracy and computational efficiency. The parameters $a$ and $b$ are calculated by feeding the data using least mean square error.

The data collected in the training sessions was fed to equations 1 and 2 without any further filtering and the final developed decoder was employed to test and control the cursor on the monitor. The upper part of Fig. 1 shows a simple schematic of this procedure.

*C. Robot interface design*

Figure 1 shows a schematic of the proposed neurofeedback-based human-robot-interaction platform. The decoded brain activity signals collected from the previous cursor control experiment are used in the offline mode to control the movements of a social robot. Here, an affordable social robot called "Rapiro" [40] is chosen to be controlled by controlled cursor position data. An Arduino and a Raspberry Pi board placed in the robot enabled us to make communication with the robot and send the command signals from the PC through Simulink [41]. Rapiro robot is a humanoid robot kit with 12 servo motors with an Arduino compatible controller board. Its capabilities for performing and controlling multitask can be extended by employing a Raspberry Pi board assembled in the head of the robot. Rapiro was selected to provide neurofeedback by executing movements, playing sounds, and flicking lights corresponding to specific commands which are extracted from decoding EEG signals. The Simulink program was compatible with making communication with the social robot and it coped with sending commands to the social robot. Here, it was programmed such that if the controlled cursor position (which was fed offline to the robot) was positive, the social robot showed right hand movement as neurofeedback for the subject; if the value was negative, the left hand movement will be the neurofeedback from the robot for the subject.

## III. RESULTS

As mentioned in the decoding section, we used five points in EEG memory data to provide a more accurate estimation for parameters of imagined body kinematics. As an example, Fig. 2 shows a plot of results from a subject during the horizontal movement training phase. It illustrates a good match between the observed cursor velocity (real values) and decoded velocity from subject's collected EEG data using the regression model. Meanwhile, Table 1 shows the results for all 5 subjects during the control of the cursor in the test phase. Four subjects each conducted 6 trials of vertical movement and 6 trials of horizontal movement. One subject conducted 6 trials of vertical movement and did not conduct horizontal movements. The total success rate of hitting the appeared random targets shows higher accuracy in horizontal movement compared to vertical movement. The subjects also reported that it was easier for them to hit the targets in the horizontal direction. This result is inconsistent with the results in other literature [11]. Here, the one dimensional movement was employed to test the developed platform in offline mode. The two dimensional movement and real-time control will be the next steps in research.

After performing the test phase by the subjects in the cursor control application, the recorded data (cursor position) for this phase are collected and they are applied to control the movements of the social robot. Figure 3 shows a series of recorded data of cursor positions that was sent in the offline mode to the Simulink to control the different parts of the social robot (e.g. right hand and left hand of social robot). Figure 3 illustrates the cursor position controlled by a subject during horizontal trials. Center of the screen, where the cursor started to move, is located at the origin (0, 0). Positive values indicate the controlled cursor is on the right side of the screen and negative values show the cursor is on the left side of the screen. After a pre-run time, the trials began and RT (Right Target) or LT (Left Target) appeared on screen. The subject had a limited time (15s) to hit the targets or the next trial would begin. In this run, the subject hit all the targets and as it is shown in Fig. 3, in all 6 trials the subject moved the cursor to the right side (positive values) for RT and left side (negative values) for LT. The subjects showed a satisfactory performance in control of the cursor during the trials except for some fluctuation at the beginning of each trial, in which the subject is managing to guide the cursor in the correct direction corresponding to the appeared target. This fluctuation is clear at the first trial in which the subject first went to the opposite direction (negative values) and then guided the cursor to the correct direction (positive values) to hit the RT.

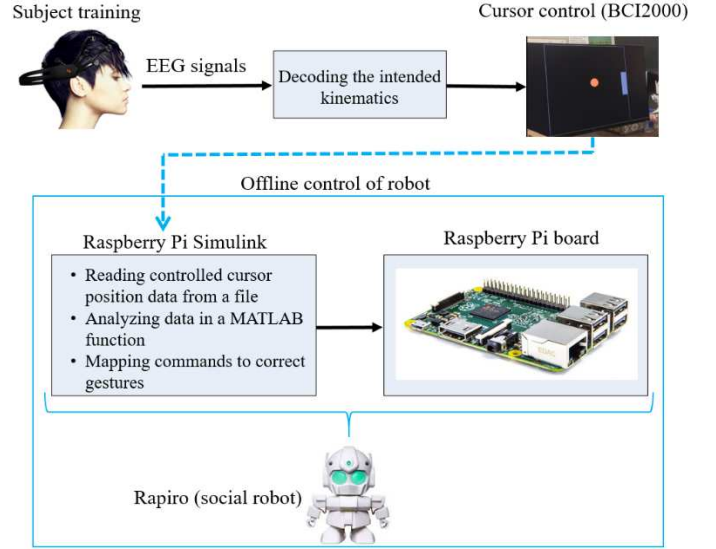

Fig. 1. A schematic of neurofeedback-based BCI platform by engaging human-robot interaction in offline mode.

Table. 1. Results of cursor movement using imagined body kinematics.

|  | Vertical Direction | Horizontal Direction |
|---|---|---|
| Number of Trials | 30 | 24 |
| Success Rate (standard deviation) | 83.3% (+/- 11.7%) | 100% (+/- 0%) |

The recorded cursor position data was fed to Simulink to control the movements of the social robot in the offline mode. As a simple experiment, it was programmed such that the social robot showed right hand movement for positive values of controlled cursor position and left hand movement for negative values of controlled cursor position. Fig. 3 shows the experimental results. In the beginning of each trial, there was a short period of time during which the robot showed incorrect hand movement, but the robot movement was quickly corrected and thereafter remained consistent with the user's intentions.

These results confirmed the validity of a platform that can be used to provide real-time neurofeedback for the subject. Here, we controlled the social robot in an offline mode. In the next step of our work, we will make direct interaction between subject and social robot and as a result, provide direct neurofeedback from the robot for the subject.

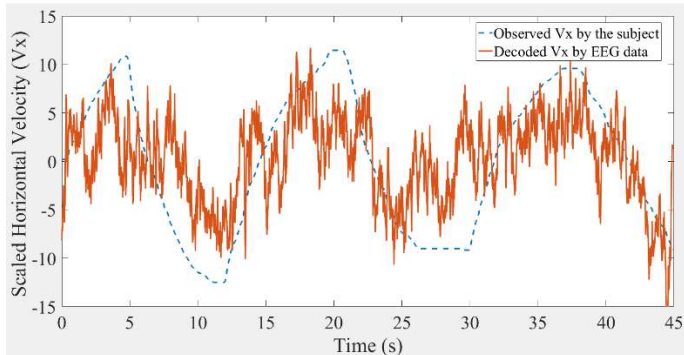

Fig. 2. Observed cursor velocity during horizontal movement training and estimated/decoded values from EEG signals by employing regression model

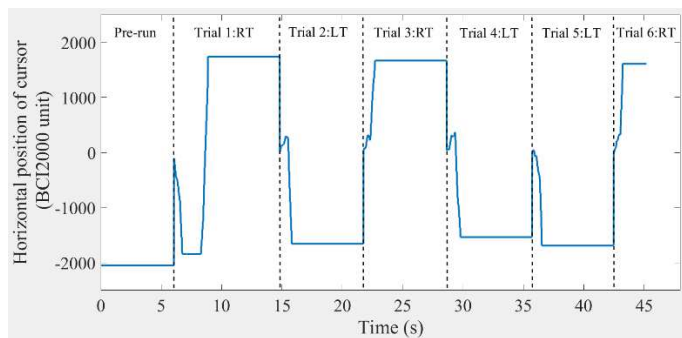

Fig. 3. Recorded values of controlled-cursor position during one run (6 trials) of cursor control in horizontal direction by a subject. RT: Right Target appeared. LT: Left Target appeared.

## IV. CONCLUSION

Brain-robot interaction has become of interest in recent years and many studies demonstrated robotic control using invasive or noninvasive brain signals. Here, as a pilot study, we presented a novel neurofeedback-based BCI platform as a testbed for cognitive training for the patient with cognitive deficits. The proposed platform is designed based on a human-robot interaction approach. For initial testing of platform, a new EEG paradigm based on continuous decoding of imagined body kinematics was used. The BCI paradigm was first applied in a computer cursor control experiment, which showed high rate of success in one-dimension of cursor control. Then, the controlled data from the cursor control task was fed into Simulink to control right hand and left hand movements of our social robot in the developed platform. The work here serves as a feasibility study to confirm the applicability of the platform for possible future development and testing with cognitive algorithms and by patients. In the future, the system will be integrated with neurofeedback exercises to improve cognitive training for patients of cognitive disorders [42-45]. Interestingly, we note that the cursor control tasks have higher accuracy in horizontal directions than in vertical directions. The discrepancy between the accuracy in horizontal and vertical controllability probably bears psychological and behavioral significance and is worthy of further investigation in future studies. One hypothesis is that horizontal eye movement may be easier than vertical eye movement and thus affect the cursor movement task correspondingly. While Bradberry et al. showed the cursor movement tasks are not the results of eye movement, there may exist secondary effects due to eye movement.

ACKNOWLEDGMENTS

This work was in part supported by a NeuroNET seed grant from University of Tennessee to XZ and by a Department of Defense grant USUHS HU0001-11-1-0007 to YJ.